\documentclass{bmvc2k}
\usepackage{multirow}
\usepackage{graphicx}
\usepackage{enumitem}
\usepackage{amsfonts}
\usepackage{bm}
\usepackage{float} 
\usepackage{subfigure}
\usepackage{amsfonts}
\usepackage{amssymb}
\usepackage{hyperref}

\title{PlaneRecNet: Multi-Task Learning with Cross-Task Consistency for Piece-Wise Plane Detection and Reconstruction from a Single RGB Image}

\addauthor{Yaxu Xie}{yaxu.xie@dkfi.de}{1}
\addauthor{Fangwen Shu}{fangwen.shu@dkfi.de}{1}
\addauthor{Jason Rambach}{jason_raphael.rambach@dfki.de}{1}
\addauthor{Alain Pagani}{alain.pagani@dkfi.de}{1}
\addauthor{Didier Stricker}{didier.stricker@dkfi.de}{1}

\addinstitution{
 German Research Centre for Artificial Intelligence (DFKI)\\
 Kaiserslautern, Germany
}

\runninghead{Y. Xie, F. Shu, J. Rambach, A. Pagani, D. Stricker}{PlaneRecNet}

\def\etal{\emph{et al}\bmvaOneDot}

\begin{document}

\maketitle

\begin{abstract}

Piece-wise 3D planar reconstruction provides holistic scene understanding of man-made environments, especially for indoor scenarios. Most recent approaches focused on improving the segmentation and reconstruction results by introducing advanced network architectures but overlooked the dual characteristics of piece-wise planes as objects and geometric models. Different from other existing approaches, we start from enforcing cross-task consistency for our multi-task convolutional neural network, PlaneRecNet, which integrates a single-stage instance segmentation network for piece-wise planar segmentation and a depth decoder to reconstruct the scene from a single RGB image.  To achieve this, we introduce several novel loss functions (geometric constraint) that jointly improve the accuracy of piece-wise planar segmentation and depth estimation. Meanwhile, a novel Plane Prior Attention module is used to guide depth estimation with the awareness of plane instances. Exhaustive experiments are conducted in this work to validate the effectiveness and efficiency of our method. \href{github.com/EryiXie/PlaneRecNet}{github.com/EryiXie/PlaneRecNet}

\end{abstract}


\section{Introduction}
\label{sec:intro}

Man-made environments usually contain rich planar surfaces. 
In indoor environments, orthogonal or parallel planar regions (walls, floor and ceiling) can form a layout that satisfies the Manhattan World (MW) assumption. Some SLAM systems~\cite{yunus2021manhattanslam, li2020structure, concha2014manhattan, flint2011manhattan, li2020rgb, rambach2019slamcraft, shu2021visual} employ such structural features for fast dense map reconstruction or low drift localization. On the other hand, planar surface representations are also utilized to solve the occlusion-awareness in indoor scene understanding~\cite{jiang2020peek}. In outdoor scenarios, an accurate ground plane estimation~\cite{man2019groundnet, rangesh2020ground} is an important prior for object pose estimation and autonomous driving. Overall, detection of 3D planes and scene reconstructing based on piece-wise planes have a very wide range of applications.

With learned prior about the geometry definition of the plane, human vision can easily detect planar regions in real world and have rough estimation of the plane normal no matter if the surface has complex texture or barely any texture, which is a difficult challenge for traditional computer vision methods. Recent deep learning approaches~\cite{Liu18-1, Liu18-2, Yang18, yu2019single} formulate the piece-wise plane estimation and reconstruction as a multi-task learning problem. Neural networks with separate branches provide predictions for segmentation mask, plane parameters and dense depth maps. The final scene reconstruction is then given by assembling the depth estimation of non-planar region and the predicted 3D planes. 

However, this assembly may result in inconsistency of the reconstructed scene between planar and non-planar regions whenever plane parameter estimation or the plane segmentation is inaccurate. Furthermore, a loss of local details on the planar region is also inevitable. Addressing this, refinement methods based on multi-view consistency~\cite{Liu18-2, xi2019reconstructing} and inter-plane relationship~\cite{qian2020learning} have been proposed to jointly improve the consistency across segmentation, parameter estimation and depth reconstruction. The geometric attributes of planes are also noticed and utilized by~\cite{Yang18}, so that the network can be trained under indirect supervision without giving piece-wise planar segmentation as ground truth. In addition to the non-optimal assembly as mentioned, we have also observed a common failure mode of existing methods in plane masks segmentation: the predicted mask can either cover multiple planes instance due to the texture similarity of these planes, or cover only a part of a plane instance due to the texture variety of that plane surface.

In this work, we propose PlaneRecNet, a multi-task convolutional neural network with a single-stage instance segmentation branch, a depth decoder with Plane Prior Attention and a shared backbone for piece-wise planar regions estimation and reconstruction. We formulate geometric constraint loss functions specific for 3D planes so that 1) the segmentation network can provide more accurate plane masks using learned occlusion clues, 2) the depth estimation in planar region is regularized with local surface normal. To summarize, our key \textbf{contributions} include:
\begin{itemize}[leftmargin=*, noitemsep]
\item A multi-task convolutional neural network for piece-wise planar reconstruction with novel Plane Prior Attention module that improves the depth reconstruction with semantic priors.
\item Novel depth gradient segmentation and plane surface normal loss functions, which enforce cross-task consistency between depth estimation and instance segmentation.
\item Extensive experiments on datasets NYUv2~\cite{Silberman:ECCV12nyu}, ScanNet~\cite{dai2017scannet} and iBims-1~\cite{koch2018evaluation}, validating the effectiveness of our method on both plane segmentation and depth estimation.
\end{itemize}



\section{Related Work}
\label{sec:related}
\noindent \textbf{Piece-wise Planar Segmentation and Reconstruction:}
PlaneNet~\cite{Liu18-1} is the first multi-task deep neural network for piece-wise planar reconstruction from a single RGB image. It consists of three prediction branches for plane parameters estimation, plane segmentation and non-planar depth map estimation with a shared encoder. 
In PlaneRecover~\cite{Yang18}, Yang and Zhou introduced a novel plane structure-induced loss to train the plane segmentation and the plane parameters estimation together through the supervision on the depth recovered from predicted planar regions and parameters, instead of ground truth mask of plane regions. 
From the aspect of segmentation accuracy, improvement was made by Z. Yu et al.~\cite{yu2019single}, who first performed semantic segmentation to distinguish planar/non-planar regions, and then clustered planar pixels into piece-wise instances with their associative embedding vectors. 
PlaneRCNN~\cite{Liu18-2} proposed a more effective plane segmentation branch built upon Mask R-CNN~\cite{he2017mask} and jointly refined the segmentation with their novel warping loss function. Qian and Furukawa~\cite{qian2020learning} proposed a post-processing refinement network to optimize the predicted plane parameters and segmentation mask of the existing piece-wise planar reconstruction method by enforcing the inter-plane relationship. Xi and Chen~\cite{xi2019reconstructing} introduced multi-view regularization method to enforce the consistency of plane feature embedding from different views.
PlaneSegNet~\cite{xie2021planesegnet} proposed a fast single-stage instance segmentation method for planes, and improved the resolution of the predicted masks.
Despite advanced general instance segmentation network architectures improving accuracy, we argue that considering the particularity of piece-wise planar segmentation and reconstruction, the specific problem constraints are not thoroughly exploited in PlaneRecover~\cite{Yang18} and Interplane~\cite{qian2020learning}. The specific consistency constraint between the piece-wise planar segmentation and the depth reconstruction can be further explored by fully using the geometry characteristics of planes.~\\

\noindent \textbf{Geometric Constraint Aided Monocular Depth Estimation:}
Recently, the effectiveness of geometric constraints in monocular depth estimation was exploited to improve the accuracy of reconstructed scene, where the point-wise loss summation over all pixels shows limitations. Novel loss functions based on geometric constraints such as the surface normal and the occlusion boundary were introduced in several approaches~\cite{ummenhofer2017demon, Yin_2019_ICCV, huynh2020guiding, long2020occlusion, yin2020learning}. 
Another open question is the proper method to obtain the reference of the geometry constraint from noisy ground truth depth, since many real datasets are obtained with consumer-level devices. 
Addressing this, Yin \etal~\cite{Yin_2019_ICCV} proposed Virtual Normal which shows robustness against local noise in experiments, thanks to the long-range triplet points sampling strategy. Long \etal~\cite{long2021adaptive} introduced an Adaptive Surface Normal Constraint. In their work, a series of random triplets are sampled within a local patch and adaptively weighted with the guidance feature given by another decoder branch in order to provide confident normal on sharp corner and edges.
Piece-wise planes are also utilized to formulate geometry constraints, especially for reconstructing indoor scenes. Huynh \etal~\cite{huynh2020guiding} indicated that the depth values of all points belonging to the same plane are linearly dependent, and they introduced a Depth-Attention Volume network to implicitly learn attention map based on co-planar relationships of each pixel.
Long \etal~\cite{long2020occlusion} introduced the Combined Normal Map, which combined the local surface normal of non-planar regions and the mean of the surface normal in a planar region
, in order to improve the reconstruction for both local high-curvature and global planar regions.
The performance of self-supervised depth estimator in indoor datasets are limited, because large non-texture planar regions are difficult to be learned only with the multi-view photometric consistency. Yu \etal~\cite{IndoorSfMLearner} utilized piece-wise planar regions annotated using superpixels as priors, to enforce a low plane-fitting error within in each planar region.~\\


\noindent \textbf{Cross-Task Consistency for Multi-Task Learning:}
As aforementioned, some works opted to train depth estimation networks with novel loss functions rather than introducing one or more separate decoders, while other works~\cite{wang2016surge, qi2018geonet, qi2020geonet++, man2019groundnet, ramamonjisoa2019sharpnet} focused on explicitly predicting surface normal, depth and other dense maps in separate streams, then jointly regulating the consistency between them and refining the initial scene reconstruction. The consistency across different tasks were not only noticed in 3D reconstruction but also in multi-task learning pairs such as depth estimation + semantic segmentation~\cite{ramirez2018geometry}, depth estimation + scenes parsing~\cite{xu2018pad}. Meanwhile some fundamental questions about multi-task learning in computer vision are also explored. Stanley \etal~\cite{standley2020tasks} gave an empirical study of influence factors of multi-task learning and a detailed investigation about how tasks influence one another. Zamir \etal ~\cite{zamir2020robust} proposed a general concept to simplify the supervision of the cross-task consistency in multi-task learning, so that the consistency constraint can be learned from data rather than a prior given relationship. So far, most of the works focus on the cross-task consistency between dense maps. Hence, we consider our work as an attempt to tackle the cross-task consistency problem between instance segmentation and dense map prediction.


\section{Approach}
\label{sec:approach}
Given a single color image $I$ as input, we use a multi-branch neural network with a shared backbone to predict piece-wise planar segmentation $M_{pred}$ and global depth estimation $D_{pred}$. We adapt a light-weight configuration of SOLO V2~\cite{wang2020solov2} for piece-wise planar segmentation, which is a global mask based single-stage instance segmentation method. Segmentation mask is provided through the dynamic convolution operation between predicted mask kernels from prediction head and mask feature from mask head. Mask candidates are then fused into the depth branch as attention through the Plane Prior Attention module. The depth decoder is a lightweight Feature Pyramid Network~\cite{lin2017feature} structure: feature maps from the encoder are upsampled through a series of convolution layers and fused with the corresponding feature maps from the encoder through skip connections.

\begin{figure}[htb]
    \centering
    \includegraphics[width=0.95\linewidth]{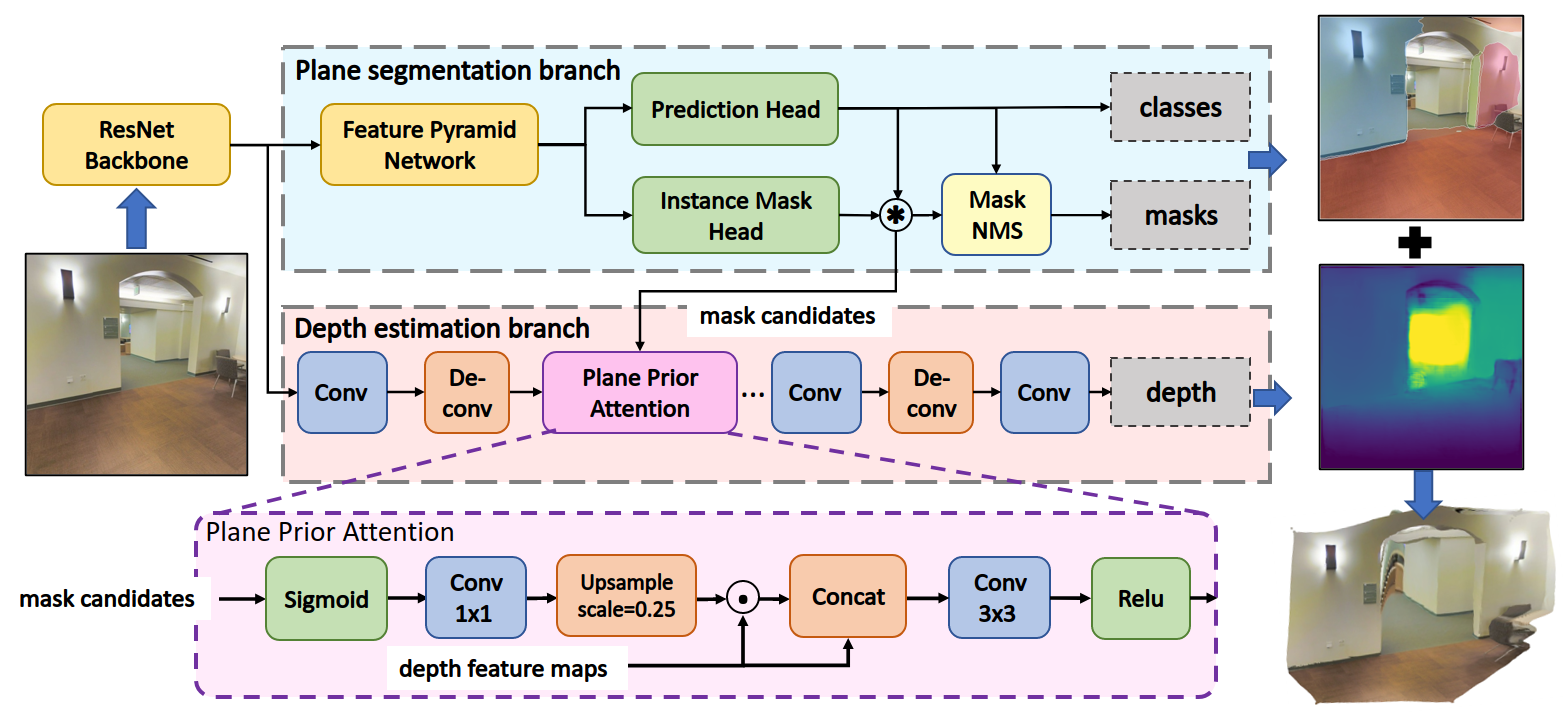}
    \caption{\textbf{The PlaneRecNet Architecture:} the network consists of a shared backbone and two task branches for piece-wise plane segmentation and monocular depth estimation. "$\odot$" presents the element-wise multiplication, "$\circledast$" presents the dynamic convolution operation.}
    \label{fig:network}
\end{figure}

Comparing to PlaneRCNN~\cite{Liu18-2}, another detection based approach for piece-wise plane reconstruction, our approach is different in following aspects: 1) We utilizes a global-mask based instance segmentation method~\cite{wang2020solov2}, which provides segmentation masks with higher resolution than local-mask based method~\cite{he2017mask}. 2) Instead of explicitly predicting plane parameters and complicating the network architecture, we focus on providing a high quality global depth estimation, so that plane parameter can be computed simply with Principal component analysis (PCA)~\cite{pca} or Random sample consensus (RANSAC)~\cite{ransac} algorithm with the plane segmentation and the depth prediction. 3) PlaneRCNN introduced an extra segmentation refinement network using global depth, rough segmentation mask and coordinate map of the plane instance as inputs. Our network has no refinement module and it is therefore simpler and faster. The network architecture of our PlaneRecNet is illustrated in Figure~\ref{fig:network}.

\subsection{Plane Prior Attention}\label{sec:ppa}
Inspired by the Depth Attention Volume~\cite{huynh2020guiding}, we design the Plane Prior Attention (PPA) module, which fuses the plane prior information to the depth decoder, as illustrated in Figure~\ref{fig:network}. We implement the Plane Prior Attention module by first introducing mask candidates for piece-wise planes into the depth branch. We then reduce the redundant channel numbers of the mask candidates using $1\times1$ convolution, and resize the mask candidates with interpolation. After that, the mask candidates (as attention weights) are multiplied and concatenated with the depth feature maps. Finally the feature maps are fed into a standard $3\times3$ convolution block followed with Relu activation.

\subsection{Guidance Loss Functions for Cross-Task Consistency} \label{sec:loss}
\begin{figure}[htb]
\centering
\subfigure[Graph structure of loss functions]{
\label{fig:loss_graph_1}
\includegraphics[width=0.55\textwidth]{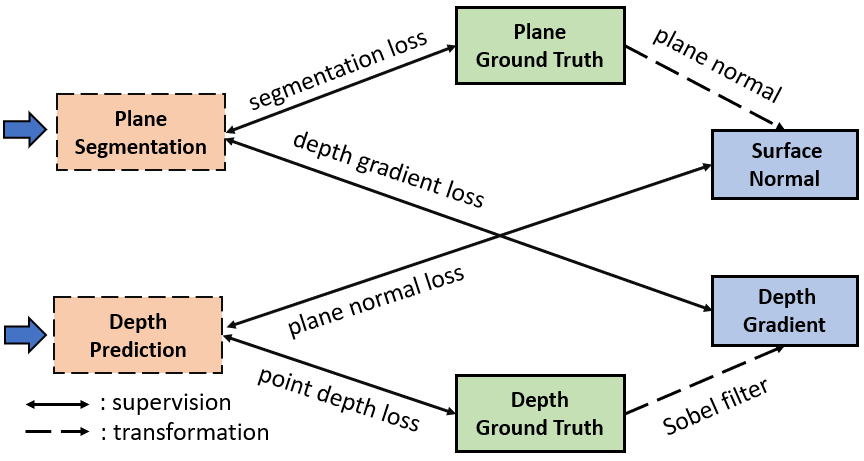}}
\subfigure[Ground truth and their conversion]{
\label{fig:loss_graph_2}
\includegraphics[width=0.32\textwidth]{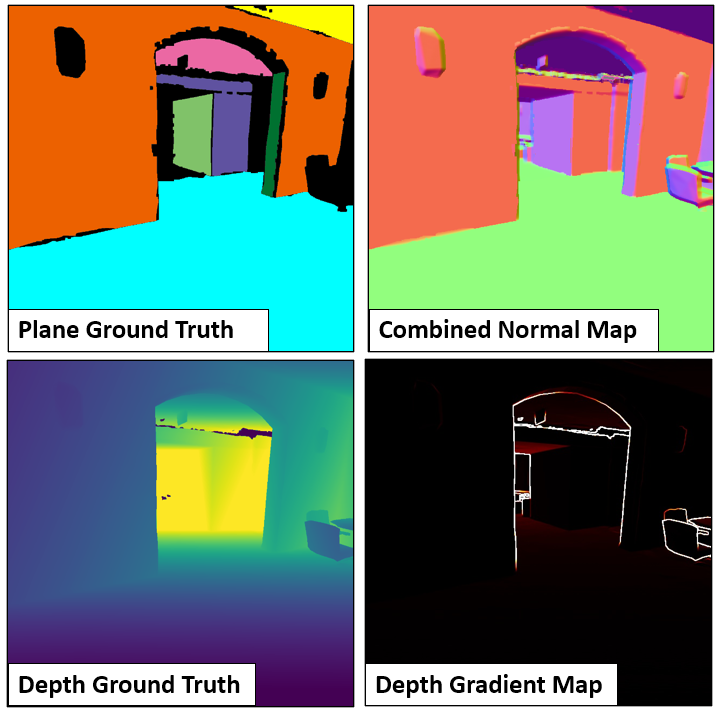}}
\caption{\textbf{Graph structure of loss functions for cross-task consistency:} with surface normal and depth gradient as intermediaries, loss functions are built into a graph, so that the plane segmentation is supervised by depth ground truth and the depth prediction is in reverse supervised by plane instance annotation.}
\label{fig:loss_graph}
\end{figure}

In piece-wise plane reconstruction, the cross-task consistency is naturally linked to the geometry of the planar surface. For depth estimation, we expect the predicted depth map can represent well the surface normal and the planarity of the planar surface of the scene. Moreover, for instance segmentation, a single predicted mask should not cover multiple planar regions, especially when these planar regions are geometrically inconsistent. To enforce such cross-task consistency, we utilize the depth ground truth as a clue of occlusion boundary to constraint the instance segmentation, and the segmentation ground truth as surface normal constraint to regularize the depth estimation, as illustrated in Figure~\ref{fig:loss_graph_1}.

\noindent\textbf{Overall Loss:} The first three base terms of the overall loss function are: Focal Loss~\cite{lin2017focal} for classification $\mathcal{L}_{Focal}$, Dice Loss~\cite{milletari2016v} for IoU based segmentation supervision $\mathcal{L}_{Dice}$ and RMSE Loss for point-wise depth supervision $\mathcal{L}_{RMSE}$. Except them, we introduce two geometry constraints, the depth gradient segmentation (DGS) loss and plane surface normal (PSN) loss, to enforce the cross-task consistency:
\begin{equation}
    \mathcal{L} = \mathcal{L}_{Focal} + \alpha\mathcal{L}_{Dice} + \beta\mathcal{L}_{RMSE} + \gamma\mathcal{L}_{DGS} + \delta\mathcal{L}_{PSN} 
\end{equation}
where $\alpha$, $\beta$, $\gamma$ and $\delta$ are weighting parameters, which are set to $\alpha = 3$, $\beta = 5$, $\gamma = 1$ and $\delta = 1$ through experiments. \\

\noindent\textbf{Depth Gradient Segmentation (DGS) Loss:}
We observe the fact that, when trained only with common segmentation loss functions based on IoU, an instance segmentation network may fail in providing the accurate mask boundary of the plane instance and have the tendency to cover a part of the other nearby plane instance, as shown in Figure~\ref{fig:ablation}. From the point of view of mask IoU based loss functions like Dice Loss, Binary Cross-Entropy~\cite{cox1958regression} or Focal Loss, such failure cases only cause a very limited loss overall and are therefore hard to learn. Especially in multiple plane regions with continuous low-texture, the segmentation performance highly relies on edges as clues. Addressing this, we propose to use the occlusion boundary of the depth map as an extra geometry constraint, because the gradient of the depth map within a planar region obviously should not have any significant changes, as illustrated in Figure~\ref{fig:loss_graph_2}. We first computed the depth gradient map $G_{gt}$ of the ground truth depth $D_{gt}$ using Sobel-Filter in $x-$ and $y-$directions, and normalize the gradient map with $D_{gt}$ for the obvious reason that far away pixels have larger gradient than nearby pixels:
\begin{equation}
    G_{gt} = \frac{G_{x}^2 + G_{y}^2}{D_{gt}^2} \quad\quad \textrm{with} \quad\quad G_{x},\  G_{y} = Sobel_{x}(D_{gt}), \ Sobel_{y}(D_{gt})
\end{equation}
With the gradient map as an intermediary between the depth ground truth and the segmentation prediction, we built the depth gradient segmentation loss (DGS Loss) simply with the element-wise multiplication of the gradient map $G_{i}$ of frame $i$ and mask candidates $P_{i,j}$ of this frame, $M$ as the number of mask candidates per frame and $N$ as the batch size:
\begin{equation}
    \mathcal{L}_{DGS} = \frac{1}{N} \sum_{i}^{N}\frac{1}{M}\sum_{j}^{M} G_{i} \odot  P_{i,j} 
\end{equation}

\noindent\textbf{Plane Surface Normal (PSN) Loss:}
Since our network is dedicated to reconstruct planes in the scene and indoor scenes are rich of planar surfaces, it is natural to constrain the depth prediction using the geometric properties of the plane. In addition to the gradient of depth, another link between plane and depth is the surface normal. 

In order to minimize the dependence on noisy depth ground truth, we adapt the Combined Normal Map proposed by Long \etal~\cite{long2020occlusion}, which literally combines the normal of planar regions and non-planar regions, as illustrated in Figure~\ref{fig:loss_graph_2}. In our implementation the normal of the planar region is assigned with the normal of the corresponding 3D plane model, rather than the mean value of the surface normal of each planar region as the original design. 

To avoid the expensive computation of estimating surface normal with differentiable least square method~\cite{qi2018geonet} during training, we utilize the random sampling strategy similar as in ~\cite{Yin_2019_ICCV, long2021adaptive}. Having pixel $p_{i}(u_{i},v_{i})$ of the predicted depth, the 3D location $P_{i}(x_{i}, y_{i}, z_{i}) \in \mathbb{R}^{3}$ can be obtained with the camera intrinsic matrix. We then divide the predicted point cloud into $N$ planar groups and one non-planar group using the plane segmentation ground truth: $P_{i} \in \mathbb{R}_{j}^{3}$, $ \mathbb{R}^{3} = \{ \mathbb{R}_{j}^{3} | j = 0,..., N\}$. We randomly sample $K$ point triplets $T_{i}=\{(P_{k}^{A}, P_{k}^{B}, P_{k}^{C}) | P \in \mathbb{R}_{j}^{3}, k = 0, ..., K-1\}$ within each group, the normal vector of a triplet is defined as:
\begin{equation}
    \vec{n_{k}} = \frac{\overrightarrow{P_{k}^{A}P_{k}^{B}} \times \overrightarrow{P_{k}^{A}P_{k}^{C}}}{\left|\overrightarrow{P_{k}^{A}P_{k}^{B}} \times \overrightarrow{P_{k}^{A}P_{k}^{C}} \right|}
\end{equation}
In~\cite{Yin_2019_ICCV}, two restrictions are introduced for each triplet: angle restriction to avoid colinearity and Euclidean distance restriction for sampling more on long-range points and being robust to local noise. 

Here we adopt two different restriction strategies for triplets within planar regions and within non-planar region. Since the normal of the triplet from predicted depth $T_{pred}$ in planar regions will be compared with the normal of the fitted 3D plane, we only restrict the triplet not to be colinear, because we expect that the predicted depth in planar regions represents the planarity both locally and globally. Triplets from predicted depth within non-planar regions are compared with their corresponding triplets from noisy ground truth depth, we therefore also restricted the Euclidean distance between triplet points. The planar guidance surface normal loss (PSN Loss) is defined with the cosine similarity function:
\begin{equation}
    \mathcal{L}_{PSN} = 1 - N_{pred}^{T}N_{gt}
\end{equation}
where $N_{pred}$ is the surface normal sample calculated from the depth prediction and $N_{gt}$ is from the Combined Normal Map with our grouping and restriction strategies.


\section{Experiments}
\label{sec:experiments}
In the section, we conduct several experiments to compare our approach against the state-of-the-art plane reconstruction methods and other monocular depth estimation approaches.

\subsection{Implementation Details}
Our proposed network, PlaneRecNet, is implemented with the Pytorch~\cite{paszke2019pytorch} framework, and trained 
using a batch size of 8 images with Adam~\cite{kingma2014adam} optimizer and a base learning rate of $1\times10^{-4}$. To stabilize the training in early stage, we set a warm-up phase to linearly increase the base learning rate from $1\times10^{-6}$ to $1\times10^{-4}$ during the first 2,000 iterations. The model is trained for 10 epochs on 10,000 samples (the same amount as the training configuration of PlaneRCNN~\cite{Liu18-2}) from ScanNet with the plane annotation given by ~\cite{Liu18-2}. We augment the training data using random photometric distortion, 
horizontal and vertical flipping 
and Gaussian noise. All experiments are conducted with the same configuration, as well as the same backbone encoder ResNet101~\cite{he2016deep} with deformable convolution~\cite{zhu2019deformable}.

\subsection{Datasets and Metrics}
To conduct the experiments, we use the labeled subset of \textbf{NYUv2} dataset~\cite{Silberman:ECCV12nyu} (the plane segmentation ground truth annotated with RANSAC), about 5,000 random sample from \textbf{ScanNet}~\cite{dai2017scannet}, and \textbf{iBims} dataset~\cite{koch2018evaluation} as benchmarks. None of these data samples is seen by our network during training. We utilize Average Precision for mask ($AP_{m}$) and bounding box ($AP_{b}$) to evaluate the plane segmentation performance. For evaluating the depth estimation, standard metrics: Absolute Relative Error ($rel$), Log 10 error ($log_{10}$), Linear Root Mean Square Error ($RMS$), Accuracy under a threshold ($\sigma_{1},\sigma_{2},\sigma_{3}$) are utilized. Details about the new metrics introduced by iBims is explained in~\cite{koch2018evaluation}.

\subsection{Ablation Study}
In this section, we conduct an ablation study to analyze the details of our approach. To validate the effectiveness of our proposed loss functions and the fusion module, we test our network by adding different components one by one and inference on NYUv2 dataset~\cite{Silberman:ECCV12nyu} and evaluate the plane segmentation\footnotemark[1] and depth prediction results. Table~\ref{tab:ablation} shows that all the components have a positive contribution to the final performance.

\begin{table*}[!htb]\begin{center}
\scalebox{0.7}{
\begin{tabular}{|l| c c c c c c |c c c c c c |}
\hline
\multirow{2}{*}{Methods}  &  \multicolumn{6}{c|}{Segmentation Metrics} & \multicolumn{6}{c|}{Depth Metrics} \\
& $AP_{m}$ & $AP_{m}^{50}$ & $AP_{m}^{75}$ & $AP_{b}$ & $AP_{b}^{50}$ & $AP_{b}^{75}$ & $rel\downarrow$ & $log_{10}\downarrow$ & $RMS\downarrow$ & $\sigma_{1}$ & $\sigma_{2}$ & $\sigma_{3}$ \\ \hline \hline
Baseline & 14.33& 36.31& 9.38& 19.92& 41.74& 16.63& 0.169& 0.070& 0.574& 0.763& 0.944& 0.986  \\\hline
Base + DGS & 15.76& \textbf{37.77}& 11.59& 20.73& \textbf{42.03}& 17.94& 0.167& 0.070& 0.575& 0.762& 0.945& 0.986  \\\hline
Base + DGS + PSN & 15.74& 37.13& 11.63& 20.95& 41.92& 18.40& 0.166& 0.070& 0.575& 0.762& 0.945& 0.986 \\\hline
\textbf{PlaneRecNet (ours)}  & \textbf{15.81}& 37.68& \textbf{11.66}& \textbf{20.96}& 41.61& \textbf{18.52}& \textbf{0.166}& \textbf{0.069}& \textbf{0.564}& \textbf{0.765}& \textbf{0.946}& \textbf{0.987} \\ \hline 
\end{tabular}}
\end{center}
\caption{Ablation studies on the contributions of proposed geometric constraints and prior module on \textbf{NYU V2} dataset~\cite{Silberman:ECCV12nyu}. Baseline: SOLO V2~\cite{wang2020solov2} with simple FPN-like depth decoder. PlaneRecNet: Baseline with PPA Module and trained with DGS and PSN losses.} 
\label{tab:ablation}\end{table*}

\begin{figure}[!htb]
\centering
\includegraphics[width=0.95\textwidth]{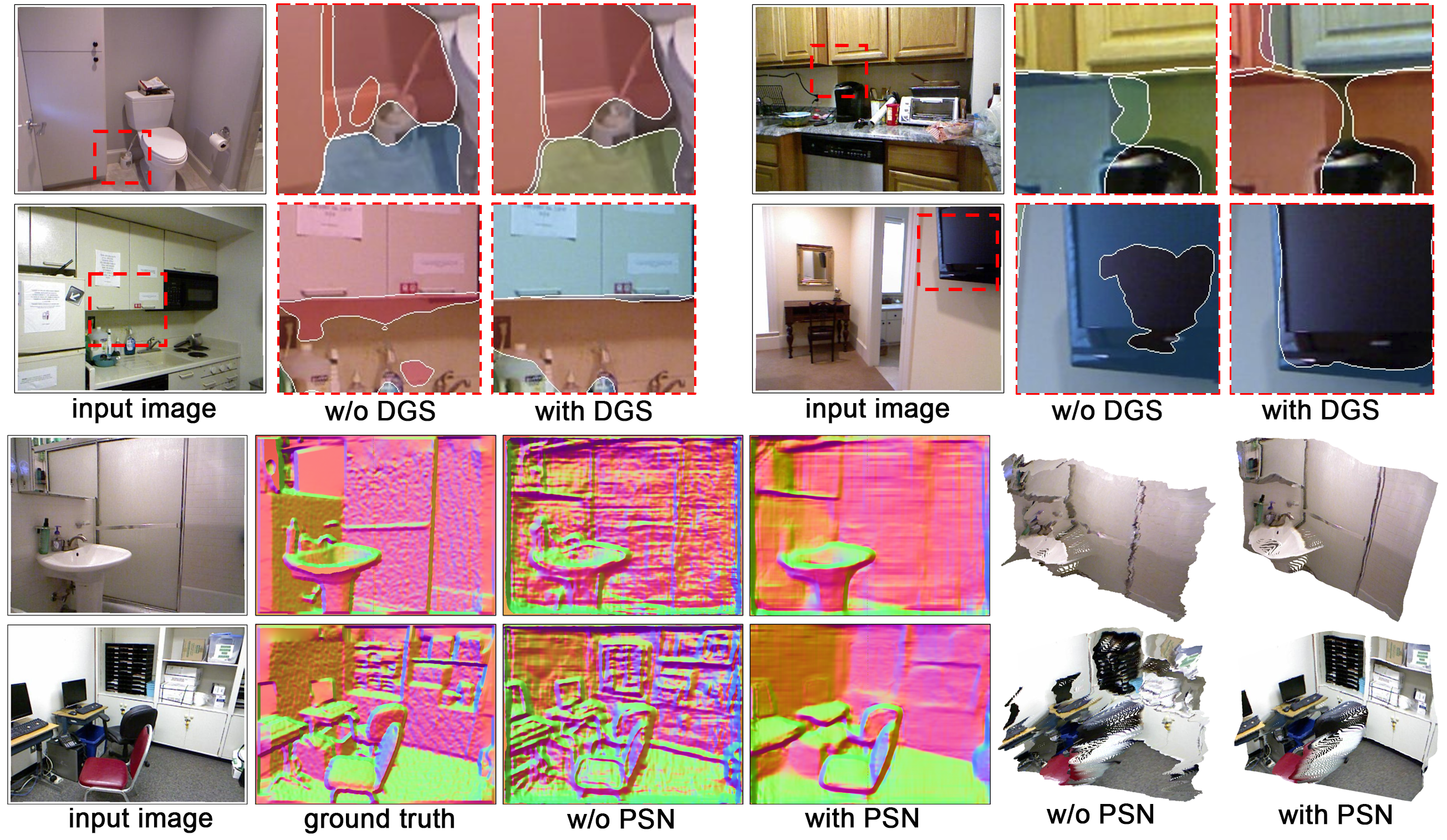}
\caption{Visualization of the effectiveness of DGS loss and PSN loss. Input images are from NYU V2 dataset.}
\label{fig:ablation}
\end{figure}

\noindent \textbf{Effectiveness of DGS:} As shown in Figure~\ref{fig:ablation}, our PlaneRecNet trained with DGS loss provides more accurate instance boundary, compare to our baseline. With depth gradient as clue, the network learns to concentrate more on occlusion boundary. The "feature leakage" problem is always a weakness of single-stage instance segmentation methods with global mask design~\cite{bolya2019yolact}. We believe that DGS loss can also be generalized to common instance segmentation task, by using instance boundaries from ground truth instead of depth gradient.\\

\noindent \textbf{Effectiveness of PSN:} With plane surface normal constraint, our depth prediction is improved in providing more precise normal map and point cloud. In Figure~\ref{fig:ablation}, we visualize the estimated surface normal using differentiable least square method~\cite{qi2018geonet} and the point cloud recovered from the predicted depth. One can see that high order geometry constraint like surface normal improves the scene reconstruction in a significant way. \\

\noindent \textbf{Effectiveness of Plane Prior Attention:}
The architecture of our PPA module is very simple. Table~\ref{tab:ablation} shows the improvement by introducing PPA module. The attention weight requires no extra supervision, since the plane segmentation mask is supervised by segmentation loss. The module can be trained along with the whole network, while Depth Attention Volume Network is trained separately in three stages~\cite{huynh2020guiding}.

\subsection{Comparison to State-of-the-art}

\begin{figure}[!htb]
\centering
\includegraphics[width=0.95\textwidth]{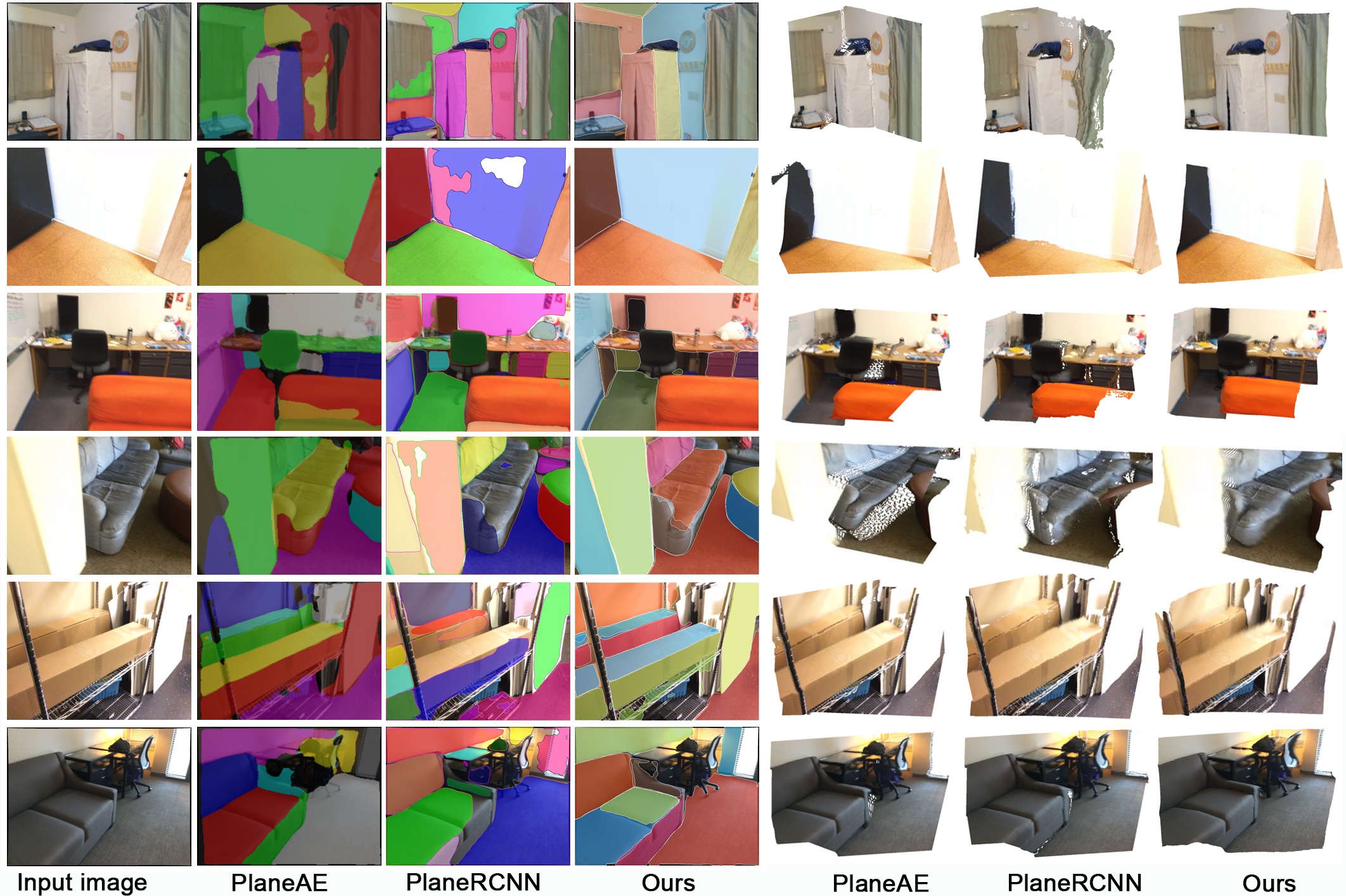}
\caption{Qualitative comparison of PlaneAE, PlaneRCNN and ours on ScanNet dataset.}
\label{fig:results}
\end{figure}

\noindent \textbf{Evaluation on ScanNet:} We evaluate our method on about 5,000 random samples from ScanNet~\cite{dai2017scannet}, and compare the results with other state-of-the-art plane reconstruction methods~\cite{yu2019single, Liu18-2} (trained on the same dataset). Since PlaneRCNN and PlaneAE are trained with SGD optimizer, we also provide the result of our network trained with it. As shown in Table~\ref{tab:seg_results}, our network outperforms other state-of-the-arts both in segmentation\footnotemark[1] and depth estimation. 
Some qualitative results are given in Figure~\ref{fig:results}. Comparing to PlaneRCNN, our method is more unlikely to counter the "uncompleted-segmentation" problem, which we believe is a side effect of the warping loss~\cite{Liu18-2} used in PlaneRCNN. The reason is that, when computing the 3D distance between the warped model and the neighbor view, a under-segmented 3D plane results less warping loss than over-segmented one. Column 1, 4 and 6 of Figure~\ref{fig:results} also show that the reconstructed point clouds from PlaneAE and PlaneRCNN have obvious inconsistency and cracks, when the plane parameters of the 3D planes are inaccurate.

\footnotetext[1]{After fixing the misimplemented metric, we updated the segmentation results, detail explained in our code.}

\begin{table*}[!htb]\begin{center}
\scalebox{0.72}{
\begin{tabular}{|l| c c c c c c |c c c c c c |}
\hline
\multirow{2}{*}{Methods}  &  \multicolumn{6}{c|}{Segmentation Metrics} & \multicolumn{6}{c|}{Depth Metrics} \\
& $AP_{m}$ & $AP_{m}^{50}$ & $AP_{m}^{75}$ & $AP_{b}$ & $AP_{b}^{50}$ & $AP_{b}^{75}$ & $rel\downarrow$ & $log_{10}\downarrow$ & $RMS\downarrow$ & $\sigma_{1}$ & $\sigma_{2}$ & $\sigma_{3}$ \\ \hline \hline
PlaneAE~\cite{yu2019single} & 5.92 & 14.72 & 4.00 & 7.86 & 17.83 & 6.25 & 0.111 & 0.049 & 0.409 & 0.864 & 0.967 & 0.991 \\\hline
PlaneRCNN~\cite{Liu18-2} & 13.38 & 26.49 & 12.23 & 14.98 & 28.54 & 13.73 & 0.124 & 0.050 & 0.265 & 0.865 & 0.972 & 0.994 \\\hline
\textbf{Ours (SGD)} & \underline{14.23} & \underline{28.23} & \underline{12.88} & \underline{17.51} & \underline{33.00} & \underline{16.00} & \underline{0.093} & \underline{0.039} & \underline{0.221} & \underline{0.917} & \underline{0.987} & \underline{0.998} \\\hline
\textbf{Ours (ADAM)} & \textbf{16.71} & \textbf{31.91} & \textbf{15.56} & \textbf{21.05} & \textbf{36.45} & \textbf{20.29} & \textbf{0.075} &\textbf{ 0.031} & \textbf{0.178} & \textbf{0.951} & \textbf{0.992} & \textbf{0.998} \\\hline 
\end{tabular}}
\end{center}
\caption{Evaluation of segmentation and depth estimation on \textbf{ScanNet} dataset~\cite{dai2017scannet}.} 
\label{tab:seg_results}\end{table*}

\begin{table*}[!htb]
\begin{center}
\scalebox{0.72}{
\begin{tabular}{|l| c c c c c c |c c| c c| c c c |}
\hline
\multirow{2}{*}{Methods} & \multicolumn{6}{c|}{Standard Metrics} &  \multicolumn{2}{c|}{PE (cm)} & \multicolumn{2}{c|}{DBE (px)} & \multicolumn{3}{c|}{DDE (\%)} \\ 
& $rel$ & $log_{10}$ & $RMS$ & $\sigma_{1}\uparrow$ & $\sigma_{2}\uparrow$ & $\sigma_{3}\uparrow$ & $\epsilon_{PE}^{plan}$ & $\epsilon_{PE}^{orie}$ & $\epsilon_{DBE}^{acc}$ & $\epsilon_{DBE}^{comp}$ & $\epsilon_{DDe}^{0}\uparrow$ & $\epsilon_{DDe}^{-}$ & $\epsilon_{DDe}^{+}$  \\\hline\hline

SharpNet~\cite{ramamonjisoa2019sharpnet}  & 0.26& 0.11& 1.07& 0.59& 0.84& 0.94& 9.95& 25.67& 3.52& 7.61& 84.03& 9.48& 6.49 \\\hline
DAV-Net~\cite{huynh2020guiding}    & 0.24& 0.10& 1.06& 0.59& 0.84& 0.94& 7.21& 18.45& 3.46& 7.43& 84.36& \textbf{6.84}& 6.27 \\\hline
VNL-Net~\cite{Yin_2019_ICCV}   & 0.24& 0.11& 1.06& 0.54& 0.84& 0.94& 5.73& 16.91& 3.65& 7.16& 82.72& 13.91& 3.36 \\\hline
\textbf{Ours}  & \textbf{0.18}& \textbf{0.09}& \textbf{1.00}& \textbf{0.67}& \textbf{0.89}& \textbf{0.95}& \textbf{3.29}& \textbf{9.12}& \textbf{2.41}& \textbf{6.59}& \textbf{84.67}& 13.91& \textbf{1.42}\\\hline \hline

PlaneNet~\cite{Liu18-1}    & 0.29& 0.17& 1.45& 0.41& 0.70& 0.86& 7.26& 17.24& 4.84& 8.86& 71.24& 28.36& 0.40\\\hline
PlaneAE~\cite{yu2019single}  $\dagger$   & 0.27& 0.15& 1.42& 0.39& 0.74& 0.88& 4.48& 10.65& 4.52& 8.39& 71.67& 27.98& \textbf{0.35} \\\hline
PlaneRCNN~\cite{Liu18-2} $\dagger$  & 0.20& 0.10& 1.04& 0.66&0.88& \textbf{0.96}& 4.50& 13.38& 4.03& 8.01& \textbf{85.51}& \textbf{12.24}& 2.25 \\\hline

\textbf{Ours + PD}   & \textbf{0.18}& \textbf{0.09}&   \textbf{1.02}&  \textbf{0.67}&   \textbf{0.89}&   0.95&   \textbf{1.55}&   \textbf{9.19}&   \textbf{3.03}&   \textbf{7.03}&  84.54& 13.92&  1.54\\\hline
\end{tabular}}
\end{center}
\caption{Comprehensive depth evaluation on \textbf{iBims-1} dataset~\cite{koch2018evaluation}. $\dagger$ Using author's released models. Ours + PD: Rendering depth map using 3D plane fitted with PCA. } 
\label{tab:dep_results}
\end{table*}

\noindent \textbf{Evaluation on iBims-1:} We evaluate our method on iBims-1 dataset~\cite{koch2018evaluation}. This dataset introduced a set of error metrics to evaluate the planarity of the depth map ($\epsilon_{PE}^{plan}$ and $\epsilon_{PE}^{orie}$), the location accuracy of depth boundaries ($\epsilon_{DBE}^{acc}$ and $\epsilon_{DBE}^{comp}$) and the consistency of depth prediction over the whole image ($\epsilon_{DDe}^{-}$, $\epsilon_{DDe}^{+}$, $\epsilon_{DDe}^{0}$). 
We provide the quantitative comparison between our method and ~\cite{ramamonjisoa2019sharpnet, huynh2020guiding, Yin_2019_ICCV, Liu18-1, yu2019single, Liu18-2} in Table~\ref{tab:dep_results}. Our network outperforms other monocular depth estimators without fitting 3D planes to the predicted depth. After fitting 3D planes to predicted depth, our networks shows better performance than PlaneNet and PlaneAE in geometry related metrics in most of the metrics, especially planarity metrics $\epsilon_{PE}^{plan}$ and $\epsilon_{PE}^{orie}$. The depth estimation accuracy of our network is actually worsen after fitting 3D planes, which confirms our argument in Section~\ref{sec:intro}, that "hard-fitting" with inaccurate segmentation masks leads to less consistent results. Additionally, our network is over \textbf{4x faster} than PlaneRCNN (runtime analysis in supplementary materials). This should be attributed to the fact that we utilize the light-weight single-stage instance segmentation method~\cite{wang2020solov2}, which is much faster than two-stage method like Mask R-CNN~\cite{he2017mask} used in PlaneRCNN. Moreover, the refinement sub-network of PlaneRCNN is also very time-consuming.


\section{Conclusion and Future Work}
\label{sec:conclusion}
This paper proposes PlaneRecNet, a multi-task convolutional neural network for piece-wise planar reconstruction from a single RGB image. We introduce depth gradient segmentation loss function and planar surface normal loss function to enforce the cross-task consistency between piece-wise plane segmentation and monocular depth estimation. An interesting future direction is to introduce a self-supervised occlusion boundary prediction branch which may further improve our network's ability of geometry understanding.

\newpage

\section{Acknowledgments}
\label{sec:acknowledgments}
The research leading to these results has been partially funded by the German BMBF project MOVEON (Funding reference number 01IS20077) and by the German BMBF project RACKET (01IW20009).
\bibliography{egbib}
\end{document}